\documentclass{ieeeaccess}
\usepackage{amsmath,amssymb,amsfonts}
\usepackage{caption}
\usepackage{stfloats}
\usepackage{cite}
\usepackage{graphicx}
\usepackage{psfrag}
\usepackage{subfigure}
\usepackage{amsmath}
\usepackage{array}
\usepackage{algpseudocode}
\usepackage{setspace}
\usepackage{blindtext}
\usepackage{url}
\usepackage{epstopdf}
\usepackage{verbatim}
\usepackage{color}
\usepackage{amsmath}
\usepackage{bm}
\usepackage{algorithm, algorithmicx}
\usepackage{multirow}
\usepackage{booktabs}
\usepackage{makecell}
\usepackage{ntheorem}
\usepackage{textcomp}
\usepackage{amsmath}
\usepackage{float}
\usepackage{soul}
\soulregister\ref7
\soulregister\cite7

\def\BibTeX{{\rm B\kern-.05em{\sc i\kern-.025em b}\kern-.08em
    T\kern-.1667em\lower.7ex\hbox{E}\kern-.125emX}}
\begin{document}
\history{Date of publication xxxx 00, 0000, date of current version xxxx 00, 0000.}
\doi{10.1109/ACCESS.2017.DOI}

\title{Tracking Based Semi-Automatic Annotation for Scene Text Videos}
\author{\uppercase{Jiajun Zhu}, \uppercase{Xiufeng Jiang},\uppercase{Zhiwei Jia},
\uppercase{Shugong Xu,
\IEEEmembership{Fellow, IEEE}},\uppercase{Shan Cao, \IEEEmembership{Member, IEEE}}}
\address{Shanghai Institute for Advanced Communication and Data Science, Shanghai University, Shanghai 200444, China (e-mail: jiajun\_zhu, xiufengjiang, zhiwei.jia, shugong, cshan@shu.edu.cn)}

\markboth
{J. Zhu \headeretal: Tracking Based Semi-Automatic Annotation for Scene Text Videos}
{J. Zhu \headeretal: Tracking Based Semi-Automatic Annotation for Scene Text Videos}

\corresp{Corresponding author: Shugong Xu (shugong@shu.edu.cn)}

\begin{abstract}
Recently, video scene text detection has received increasing attention due to its comprehensive applications. However, the lack of annotated scene text video datasets has become one of the most important problems, which hinders the development of video scene text detection. The existing scene text video datasets are not large-scale due to the expensive cost caused by manual labeling. In addition, the text instances in these datasets are too clear to be a challenge. To address the above issues, we propose a tracking based semi-automatic labeling strategy for scene text videos in this paper. We get semi-automatic scene text annotation by labeling manually for the first frame and tracking automatically for the subsequent frames, which avoid the huge cost of manual labeling. Moreover, a paired low-quality scene text video dataset named Text-RBL is proposed, consisting of raw videos, blurry videos, and low-resolution videos, labeled by the proposed convenient semi-automatic labeling strategy. Through an averaging operation and bicubic down-sampling operation over the raw videos, we can efficiently obtain blurry videos and low-resolution videos paired with raw videos separately. To verify the effectiveness of Text-RBL, we propose a baseline model combined with the text detector and tracker for video scene text detection. Moreover, a failure detection scheme is designed to alleviate the baseline model drift issue caused by complex scenes. Extensive experiments demonstrate that Text-RBL with paired low-quality videos labeled by the semi-automatic method can significantly improve the performance of the text detector in low-quality scenes.
\end{abstract}


\titlepgskip=-15pt

\maketitle

\section{Introduction}
\PARstart{S}{cene} text detection is a fundamental and important task in computer vision since it is a key step towards scene text recognition \cite{48} and many downstream text-related applications, such as text-based video understanding \cite{10} and license plate recognition \cite{49}. Benefiting from the development of convolutional neural networks (CNNs), image scene text detection has made much progress in the last few years \cite{6,7,8,36,37}, while video scene text detection has stagnated due to not only the lack of annotated video scene text dataset but also the challenges of scale variation, blur, and low-resolution in videos. 

The annotation of video text instances has always been a troublesome problem, not only to mark the location of the text instance in the current frame but also to establish the consistency of the same text instance identity(ID) between frames. Currently, almost all datasets are manually labeled \cite{5,9,10}, which is expensive and time-consuming. In addition, PaddleOCR proposes to label text instances with the text detector. However, for a large number of scene text videos to be labeled, the labeling method based on the text detector is very slow and cannot label low-quality text instances exactly. In order to reduce the cost caused by data annotation and improve the efficiency of labeling, combined with the needs of video text instances annotation, we observe that the tracking algorithm based on the correlation filter \cite{11,12,44,43} is undoubtedly a good tool to assist data annotation for scene text videos. One of the main reasons is that the text instances are relatively simple to be tracked.

In the past few years, only a few datasets \cite{1,2,3} are proposed for video scene text detection, while these datasets have two common disadvantages. One is that the number of datasets is not large enough, in which the number of scene text videos basically does not exceed 100. Another is that video scenes contained in these datasets are relatively single and the quality of datasets is good without any low-resolution, blurry, and other low-quality scenes. However, low-quality video scene text detection is a hard task that must be solved, because most of the videos in the security and surveillance fields are low-resolution and blurry. Based on the above observation and analysis, it is urgent and significant to propose a large-scale low-quality scene text video dataset with low-resolution and blurry scenes.

\begin{figure*}[t] 
\centering 
\includegraphics[scale=0.55]{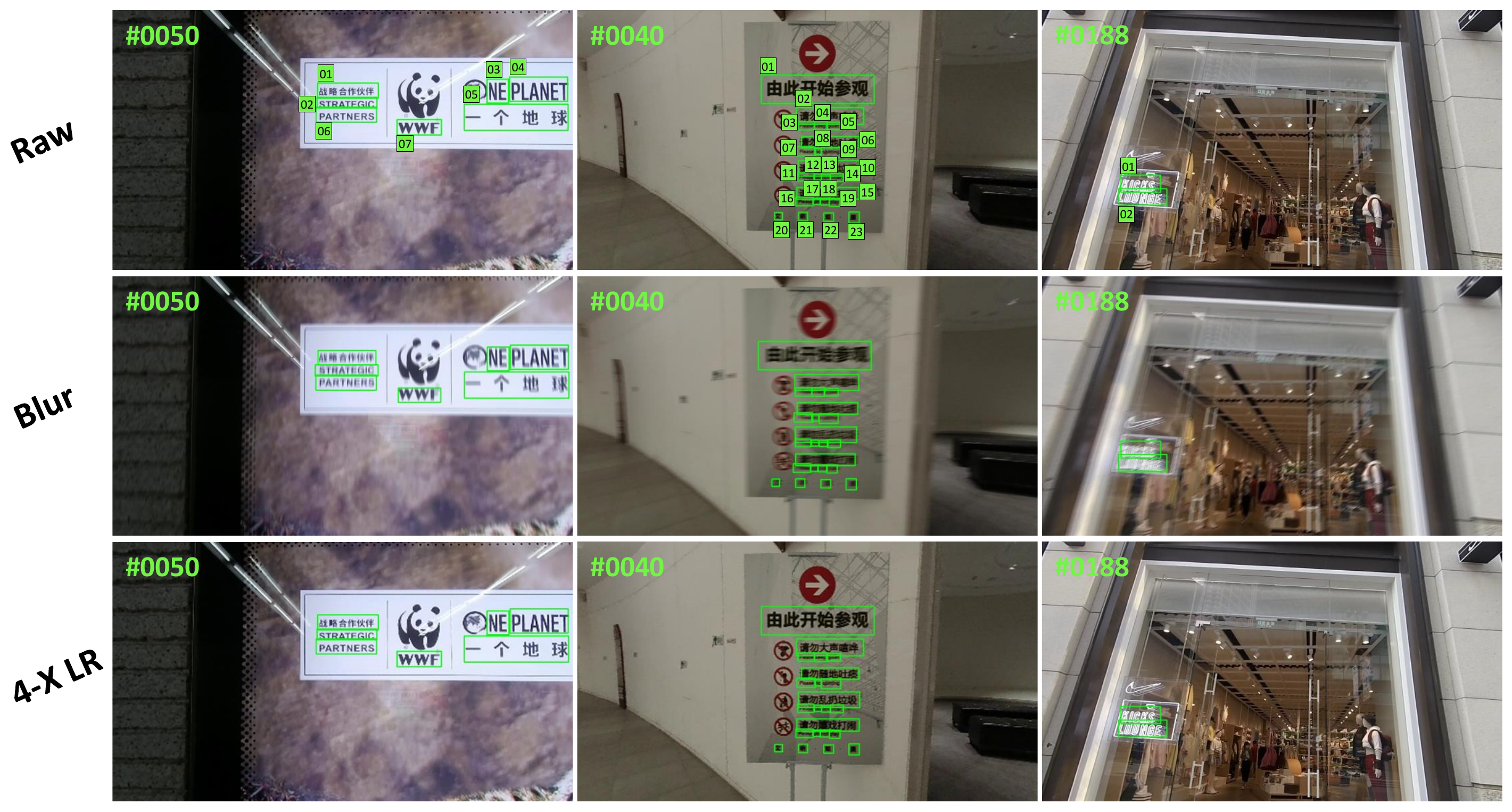}
\caption{Example paired frames and annotation of  Text-RBL. In the first row,`$Raw$' denotes the frames collected by our digital camera without any processing. In the second row, `$Blur$' denotes the paired blurry frames by the averaging operation. In the last row, `$4${-}$X$ $LR$' denotes the paired low-resolution frames by the 4 times bicubic down-sampling operation. Except for the annotations in the first frame, which are manually annotated, all annotations are the tracking results of the KCF \cite{11} tracker. In a video, we assign IDs to different text instances starting from the upper left corner, such as `$01$'. Best viewed in color and zoom in. }
\label{fig-intro} 
\end{figure*}

With the above motivations, different from the expensive and time-consuming manual labeling, a semi-automatic labeling strategy based on tracking for scene text video is proposed, which is more simple and effective. Specifically, for one scene text video, we only need to label the first frame manually. Then, for the subsequent frames, the fast tracker KCF \cite{11} based on the correlation filter is used for automatic and efficient annotation. In order to ensure that the annotation quality is good enough, manual checking and error-correcting steps are also designed. Benefiting from the proposed semi-automatic labeling strategy, not only the text instances can be annotated well in each frame, but also the same text instance have a consistent ID from frame to frame. Moreover, the low-quality scene text instances can even easily get the annotation in videos. Note, the premise of using a semi-automatic annotation strategy for scene text video annotation is that text instances appearing in the first frame will not be out of view and no new text instances appear in the subsequent frames. Therefore, for one scene text video randomly shot, we need to intercept it to obtain a sub-video that meets the requirements for annotation.

In addition, a novel and challenging scene text video dataset named Text-RBL is proposed, containing a total of 369 videos, as shown in FIGURE \ref {fig-intro}. In order to obtain richer scene text videos, our scenarios include indoors, outdoors, day, and night. What’s more, under the consideration of the videos that need to be processed in the security and surveillance fields often appearing blurry and low-resolution, we separately process the raw text videos captured by the digital camera with two different operations to obtain the paired blurry videos and low-resolution videos. Specifically, an averaging operation often mentioned in deblurring methods \cite{16,18,41,42} with a sliding window mechanism is used to generate blurry videos, and a bicubic down-sampling operation often mentioned in super-resolution methods \cite{17,19,39,40} is used to generate low-resolution videos. The annotation of Text-RBL is completed by the proposed semi-automatic labeling strategy.

In order to prove the validity of the proposed Text-RBL dataset labeled by the semi-automatic labeling strategy, a baseline model is proposed, combining the text detection and tracking into a unified framework for video scene text detection. However, it is challenging as the baseline model often drifts caused by scale variation, occlusion, and out-of-view, etc. A simple and effective failure detection scheme is designed to  address  this problem. In the experiment, we analyze the performance improvement of the baseline in low-quality scene text videos brought by the text detector trained with low-quality data in Text-RBL.

The main contributions of this work can be summarized as follows:

\begin{itemize}
	\item A semi-automatic labeling strategy for simple and efficient annotation in scene text videos. To our knowledge, we are the first one to integrate tracking to get semi-automatic scene text annotation.
	\item A paired low-quality scene text video dataset named Text-RBL, which is annotated by the semi-automatic labeling strategy. Text-RBL consists of 369 paired Raw-Blurry-LR videos with a simple averaging operation and a bicubic down-sampling strategy.
	\item A detailed analysis of the Text-RBL dataset with extensive experiments is provided. Text-RBL enhances favorably the text detector to better handle low-quality scene text detection.

\end{itemize}

The rest of this paper is organized as follows. Section \uppercase\expandafter{\romannumeral2} briefly outlines the related prior work of video scene text dataset, video scene text detection, visual object tracking. Section \uppercase\expandafter{\romannumeral3} presents the details about our proposed semi-automatic labeling strategy. We introduce Text-RBL dataset including design principle, data collection, and annotation provided by the proposed semi-automatic labeling strategy in Section \uppercase\expandafter{\romannumeral4}. In Section \uppercase\expandafter{\romannumeral5}, we introduce our baseline model used for
the Text-RBL dataset. In Section \uppercase\expandafter{\romannumeral6}, we compare our model with several
baselines on the Text-RBL dataset. We discuss future work in Section \uppercase\expandafter{\romannumeral7}
, and conclude the paper in Section \uppercase\expandafter{\romannumeral8}.

\begin{table*}[t]
\centering
\renewcommand\arraystretch{1.3}
\setlength{\tabcolsep}{5.4mm}
\smallskip\begin{tabular}{|l|l|l|l|l|}
\hline
Datasets & \# Videos & \# Frames & \# Instances & Time cost of annotation\\
\hline
Minetto \cite{1} & 5 & 3,599 & 8,706 & - \\

IC13 \cite{2} & 28 & 15,277 & 93,934 & - \\

IC15 \cite{3} & 49 & 27,824 & - & - \\

LSVTD \cite{4} & 100 & 66,700 & 569,300 & 3500 hours \\

STVText4 \cite{5} & 106 & 161,347 & 1,419,008 & 2 months \\
\hline
\textbf{Text-RBL} & \textbf{369} & \textbf{101,019} & \textbf{371,955} & \textbf{72 hours} \\
\hline
\end{tabular}
\caption{Statistic comparison between our Text-RBL dataset and other video scene text detection datasets.}
\label{t4}
\end{table*}

\section{Related Work}
In this section, we briefly review the related work of video scene text dataset, video scene text detection, visual object tracking.

\subsection{video scene text dataset}
For the task of video scene text detection, the video scene text data set is an indispensable part, which is closely related to the development of video scene text detection. However, manual labeling of video scene text is expensive and time-consuming. Since not only the location of the text instance must be marked, but the same text between different frames must have the same ID, the existing data sets are all small in data volume. For example, Minetto’s dataset \cite{1} only has five text videos. ICDAR 2013 Video Text \cite{2} provide 28 videos in total, including13 videos comprised the training set and 15 the test set. ICDAR 2015 Video Text \cite{3} is substantially updated bring the number of sequences up to 49 based on ICDAR 2013 Video Text. LSVTD \cite{4} and STVText4 \cite{5} are two relatively large-scale datasets, with 100 videos and 106 videos respectively. But the cost of manual labeling is too great, for STVText4 data annotation, Cai et al. \cite{5} invited over 200 people to conduct three rounds of cross-checking for more than two months.

In this work, we propose a semi-automatic labeling strategy, which greatly reduces the time and labor costs caused by labeling.

\subsection{video scene  text detection}
Exiting video scene text detection methods can be divided into two categories, text detection in videos and text tracking in videos. Text detection in video approaches \cite{20,21,22,23} are mainly enhancing the existing image text detector to help improve feature representation and extract text instances directly in the single frames. However, these approaches still cannot perform well due to complicated temporal characteristics, such as blur, occlusion, etc. Text tracking in video approaches \cite{24,25,26} focus on capture the temporal motion of text instances between frames. For example, Zuo et al. \cite{27} and Tian et al. \cite{28} propose to integrate tracking-by-detection based methods into the video scene text detection framework.

In this work, we propose a simple baseline for video scene text detection, which fully combines the advantages of an image scene text detector and tracker.

\subsection{visual object tracking}
There have been great interests in visual tracking recently. Exiting main kinds of visual tracking approaches can be divided into two categories as based on deep learning \cite{29,30,46,47} and correlation filter \cite{14,13,15} approaches. Benefiting from the powerful representation of deep convolutional neural networks, visual tracking approaches based on deep learning \cite{31,32,45} has made much progress in the last few years. These trackers are usually very efficient, while the online update of the deep learning model is time-consuming. The approaches based on correlation filters are known for their fast speed since the correlation filter allows for fast run-times as computations are performed in the Fourier domain \cite{11}. Its tracking performance is okay for simple objects, but its performance is not good enough for complex objects.

In this work, KCF, a fast tracker based on a correlation filter, is used to label data semi-automatically. One of the main reasons is that the text instances are relatively simple, and its annotation effect can meet our expectations.

\begin{figure*}[t] 
\centering 
\includegraphics[scale=0.57]{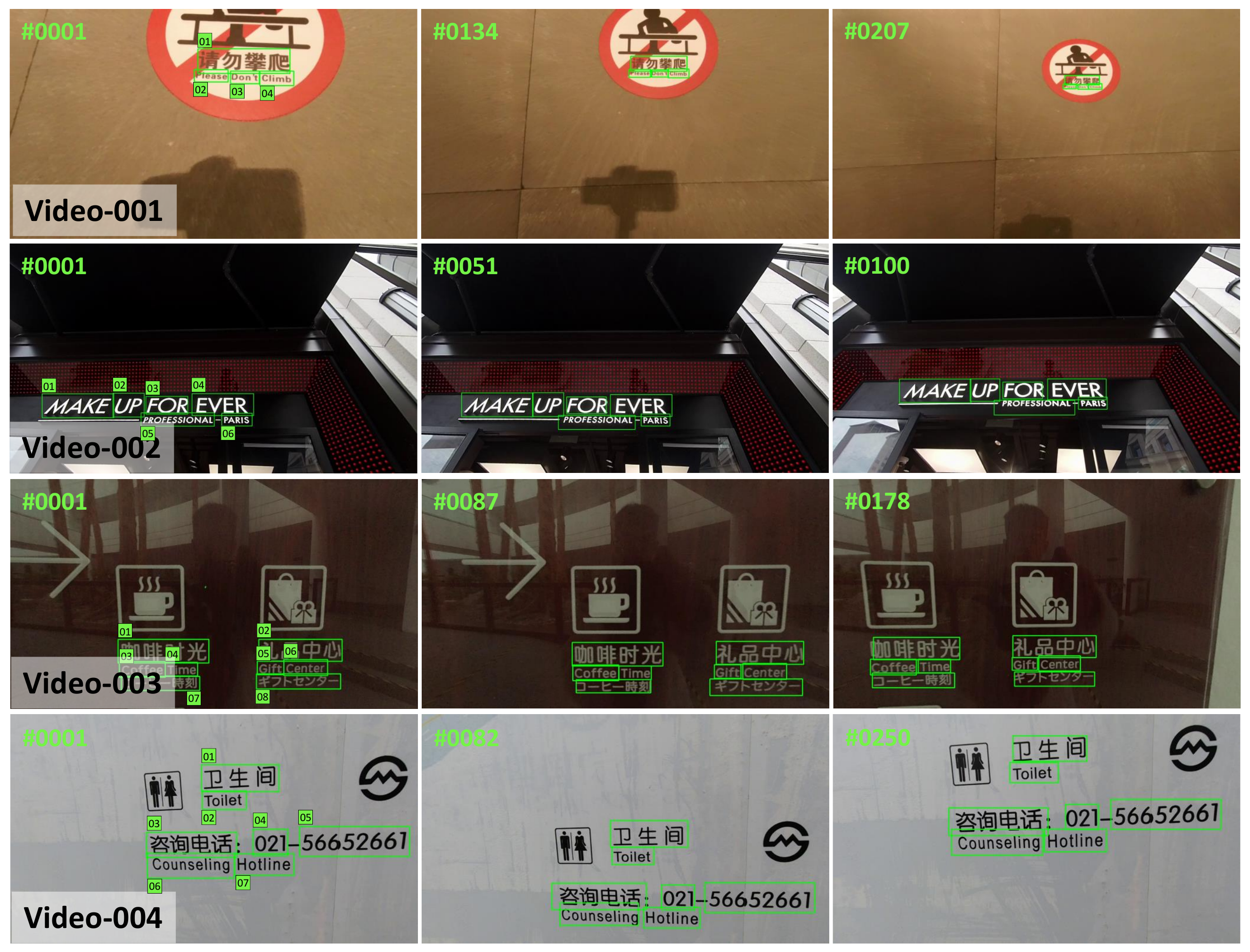}
\caption{Example raw scene text videos and annotations of our Text-RBL. The frames of each line are sampled from the same video. In each line, the first frame is the first frame of each video, which is manually labeled, and the other frames are labeled by the KCF tracker. Best viewed in color and zoom. }
\label{fig-anno} 
\end{figure*}

\section{Semi-automatic labeling strategy}
Different from the time-consuming and expensive manual labeling, in order to provide consistent bounding box annotation for scene text video efficiently, a semi-automatic labeling strategy is proposed. Under the consideration of the simple enough features of the text instance, we observe that the tracker based on correlation filter can get good enough tracking results for the text instances by using , which is one of the useful tools to obtain the temporal relation of text instances. Therefore, we propose to use the KCF tracker for automatic and efficient annotation, which is a representative, effective, and fast tracker based on the correlation filter. Note, the assumption is that in one video, except for the text instances appearing in the first frame, no new text instances appear in subsequent frames nor the old text instances disappear.

Specifically, given a scene text video, for the first frame, if a text instance appears in the frame, a labeler manually draws its bounding box as the most suitable up-right one to fit any visible part of the text instance, and the manually labeled feature of the text instance in the first frame is directly used to initialize the KCF tracker; for the subsequent frames, the initialized KCF tracker track the text instance until the entire video is completed. The final annotation of the text instance in this scene text video is combined with the result of manual annotation in the first frame and the tracking results with the KCF tracker in the subsequent frames. In fact, there is rarely only one text instance in a video, usually multiple text instances. For our semi-automatic labeling strategy, it does not increase too much workload. All we need to do is to manually label all the text instances in the first frame, and initialize a KCF tracker for each text instance in the subsequent frames to get the annotation. Moreover, we assign IDs to different text instances starting from the upper left corner in the first frame. Note that, such a semi-automatic labeling strategy cannot guarantee to get a box as accurate as manual labeling that the background area is minimized in the box. However, the strategy does provide a consistent annotation for video scene text, which is relatively stable and good enough.

Although the above semi-automatic labeling strategy is effective in most cases, there are exceptions, especially video scene text with excessive movement. In order to ensure that the annotation quality is good enough, two extra steps including manual checking and error-correcting are designed. In the manual checking process, we do not check the annotation of every text instance in each frame, which is too time-consuming, but only check the last frame of each video. If the quality of the annotation in the last frame is not good enough, we will relabel it in the error-correcting process, even manually label it if necessary, but this is an extremely rare case. FIGURE \ref{fig-anno} shows example annotation of the proposed Text-RBL dataset.

\section{The Proposed Text-RBL Dataset}
We introduce a novel video scene text dataset consisting of paired Raw-Blurry-LR videos, named as Text-RBL, detailed in Table \ref{t4}. In the following, we first design principles for constructing the Text-RBL dataset in Section \uppercase\expandafter{\romannumeral4}-A. We then provide the details of data collection, including raw videos, blurry videos, and low-resolution videos in Section \uppercase\expandafter{\romannumeral4}-B.  Following that, we introduce the automatic and efficient annotation benefiting from the proposed semi-automatic labeling strategy in Section \uppercase\expandafter{\romannumeral4}-C.

\begin{figure}[t] 
\centering 
\includegraphics[scale=0.42]{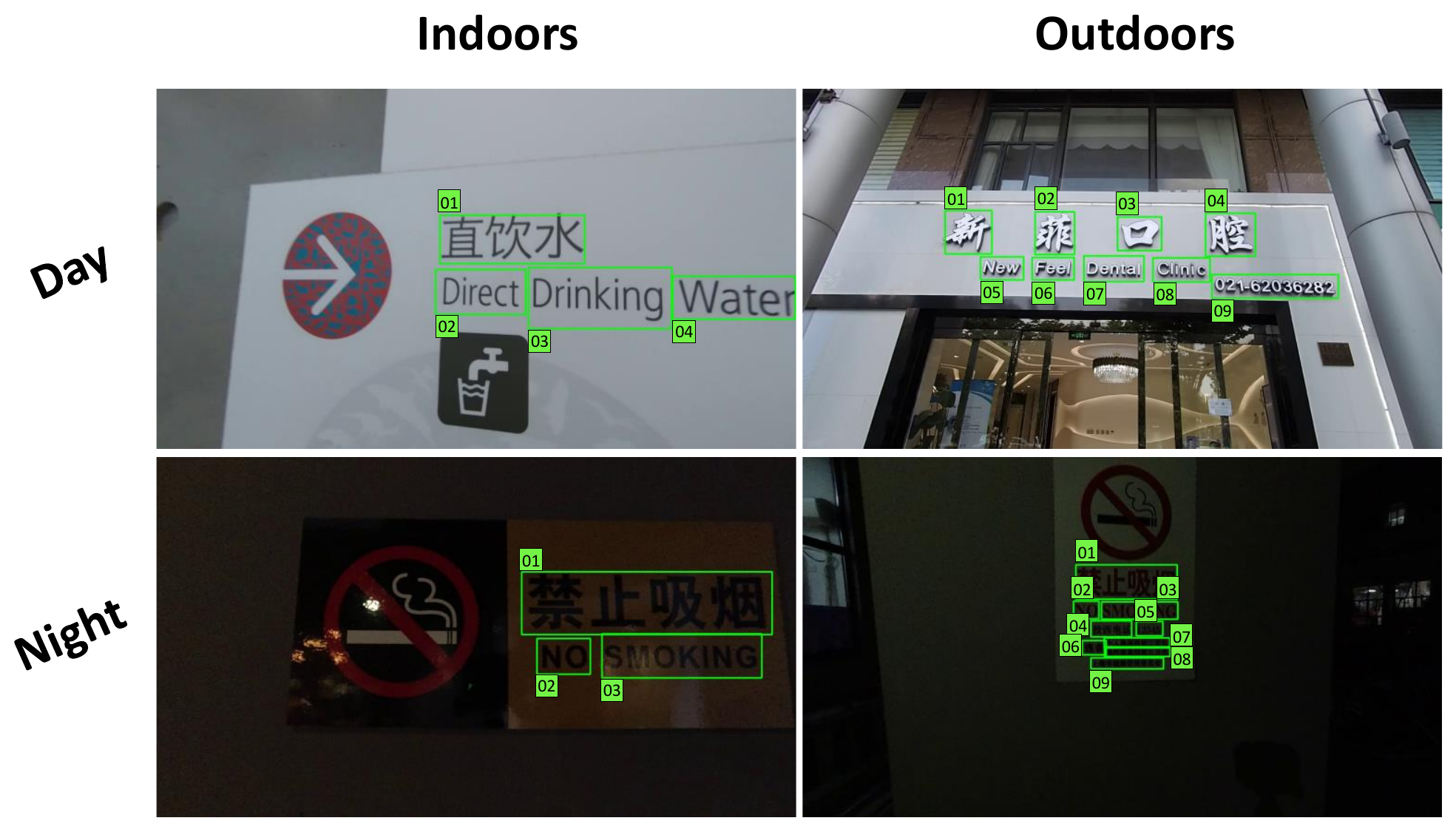}
\caption{Illustration of video text scenes in Text-RBL. Best viewed in color and zoom in.}
\label{fig-day} 
\end{figure}

\subsection{Design Principle}
Our proposed TextBlur dataset aims to offer the community a paired Raw-Blurry-LR scene test video dataset for video scene text detection. To such purpose, we follow three principles in constructing Text-RBL, including diversified scenes, efficient annotation, and comprehensive labeling.

\subsubsection{Diversified scenes}
A robust text detector is expected to perform consistently regardless of the scenes the text video belongs to. For this purpose, our video text scenes include not only indoor but also outdoor, not only during the day but also at night. Among them, indoor includes some shopping malls, museums, etc., and outdoor includes parks, streets, etc.

\subsubsection{Efficient annotation} 
The annotation of video text data has always been an expensive and time-consuming task. To alleviate this problem, we use tracking for semi-automatic and efficient annotation, which will later be discussed in Section \uppercase\expandafter{\romannumeral4}-C.

\subsubsection{Comprehensive labeling} 
In order to meet the needs of video scene text detection, a principle of Text-RBL is to provide comprehensive labeling for paired Raw-Blurry-LR scene text video, including both bounding box and text annotations. 

\subsection{Data Collection}
Text-RBL consists of paired Raw-Blurry-LR videos in diversified scenes. Different from the existing public-available video text dataset \cite{1} that only has five real outdoor videos, our video text scenes include not only outdoor but also indoor, such as streets, museums, etc, not only in the day but also at night, as shown in FIGURE \ref{fig-day}.

\begin{figure}[t] 
\centering 
\includegraphics[scale=0.37]{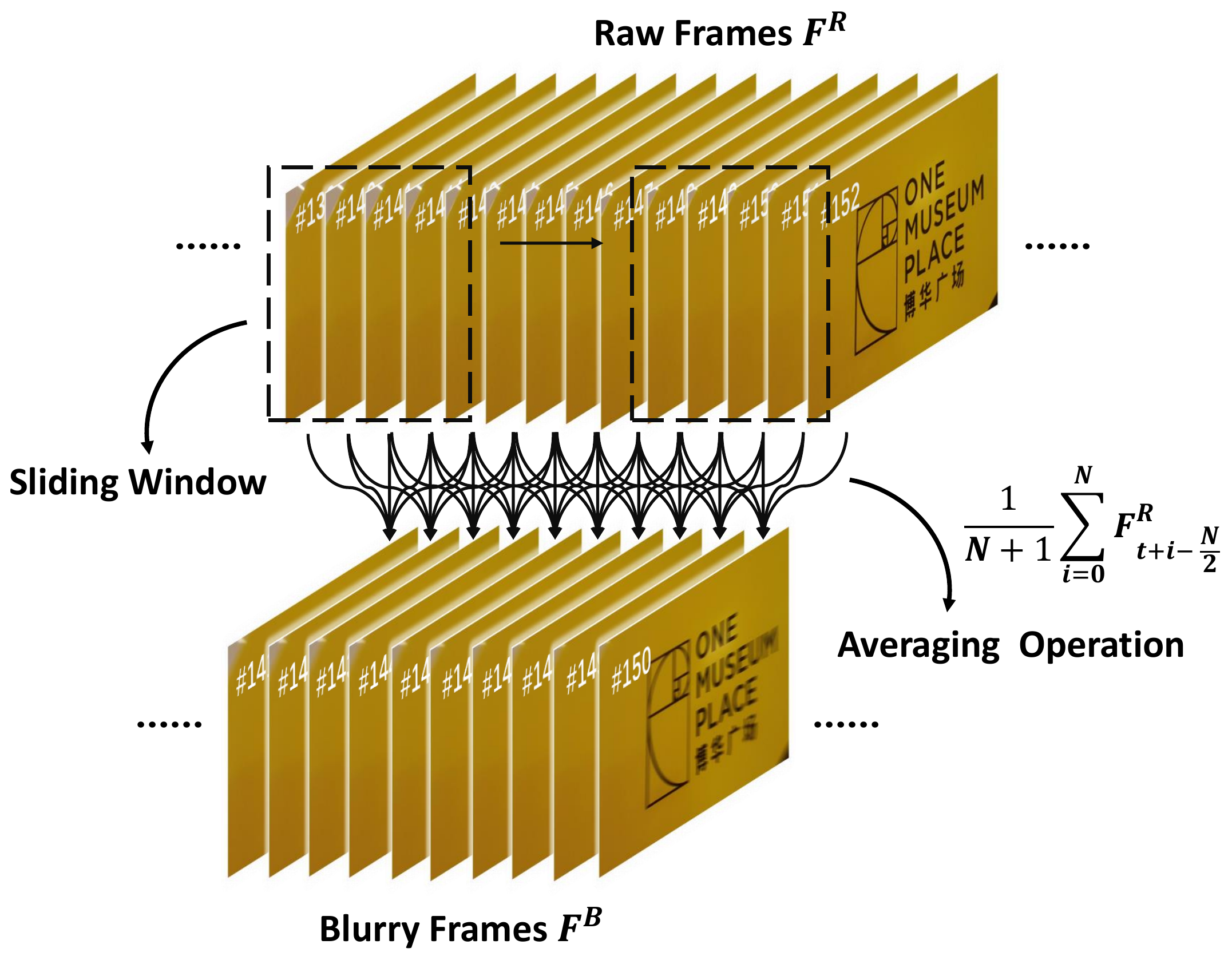}
\caption{Illustration of the proposed blurring process with an averaging operation and a sliding window to generate paired Raw-Blur videos. For the $t$-th raw frame $F_t^R$ in the video, we use the N frames before and after it with itself to generate the blurry frame $F_t^B$. We empirically set $N=5$. Best viewed in color and zoom in.}
\label{fig-blur} 
\end{figure}

\subsubsection{Raw videos}
All raw videos in Text-RBL are captured by a digital camera DJI Osmo Action with the size of 1920×1080 (width×height) and the frame rate of 100fps. Initially, we collect over 500 videos. Under the joint consideration of the quality videos for text detection and the design principles of Text-RBL, we pick out 369 videos eventually. Note, in order to use the semi-automatic labeling strategy more conveniently, when collecting videos we adhere to a principle: the text instances appearing in the first frame should be always in the view and no new text instances appear in the subsequent frames.

\subsubsection{Blurry videos}
As mentioned in conventional deblurring methods, we generate blurry frames with a simple averaging operation. The difference is that we need to generate paired Raw-Blurry videos, therefore, we additionally add a sliding window mechanism, as shown in FIGURE \ref{fig-blur}. Specifically, for the $t$-th raw frame $F_t^R$ in the video, we use the N frames before and after it combined with itself by an averaging operation to generate the blurry frame $F_t^B$. The blurring process can be mathematically formulated as:
\begin{equation}
F_t^B = \frac{1}{(N+1)}{\sum\limits_{i=0}^{N}}{F_{t+i-\frac{N}{2}}^R},
\end{equation}

\begin{figure}[t] 
\centering 
\includegraphics[scale=0.28]{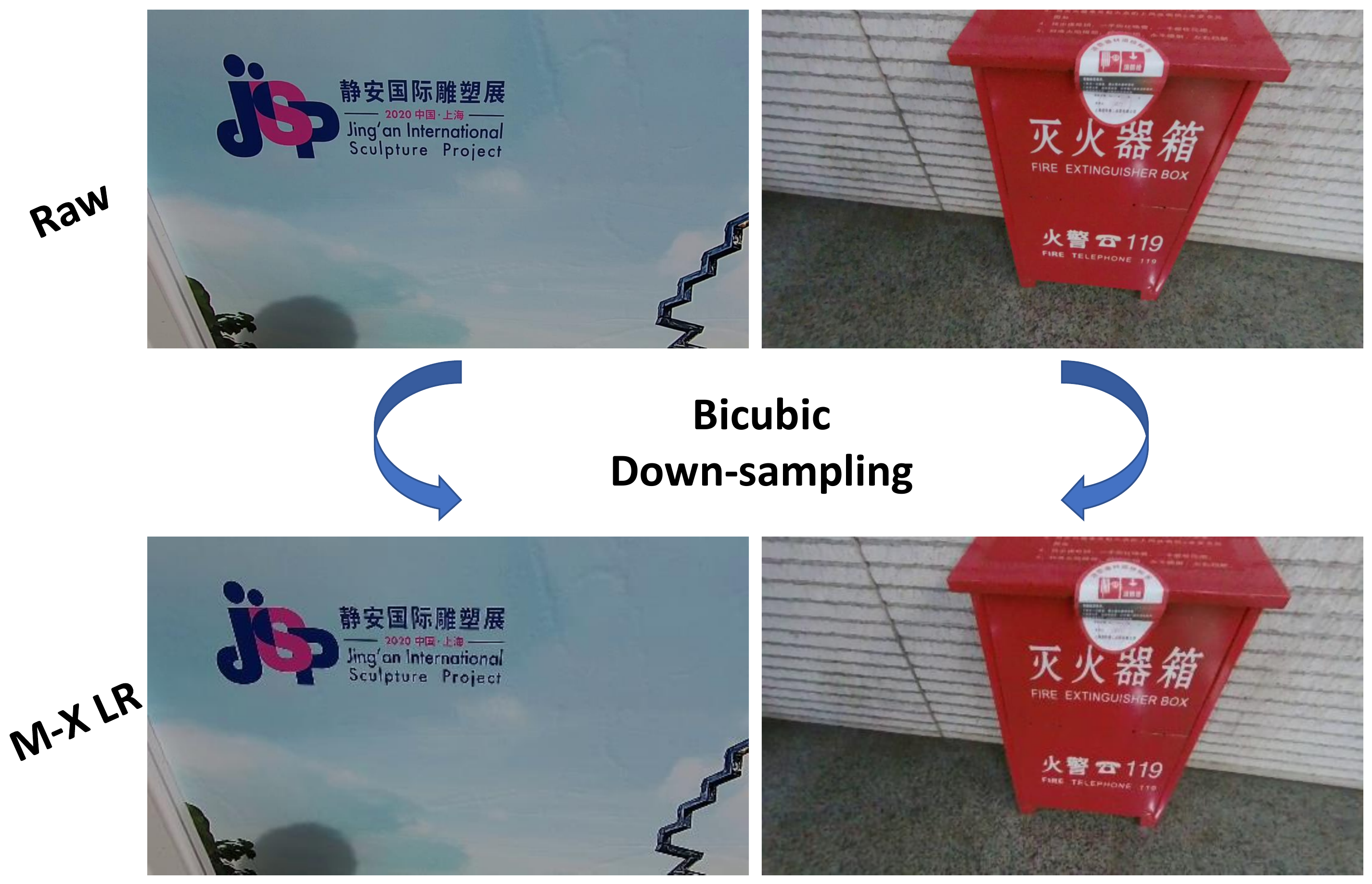}
\caption{Comparison between raw frames and LR frames in TextBlur. `$Raw$' denotes the frames collected by our digital camera. `$M${-}$X$ $LR$' denotes the low-resolution frames generated by the bicubic down-sampling strategy with multiples M . We empirically set $M=4$. Best viewed in color and zoom in. }
\label{fig-LR} 
\end{figure}

\subsubsection{Low-resolution videos}
In addition to generating paired Raw-Blurry videos, we also propose paired Raw-LR videos. As mentioned in traditional super-resolution methods, we simply use the bicubic down-sampling strategy to get low-resolution videos, as shown in FIGURE \ref{fig-LR}. Practically, through different downsampling multiples M, we can get the corresponding M-X LR videos. The LR process can be mathematically formulated as:
\begin{equation}
F_t^L = g_M(F_t^R),
\end{equation}
where $g_M(\cdot)$ is the bicubic down-sampling operation, and M is the parameter of downsampling multiple. $F_t^L$ is an M-X LR frame corresponding to the raw frame $F_t^R$.

\begin{figure}[t] 
\centering 
\includegraphics[scale=0.45]{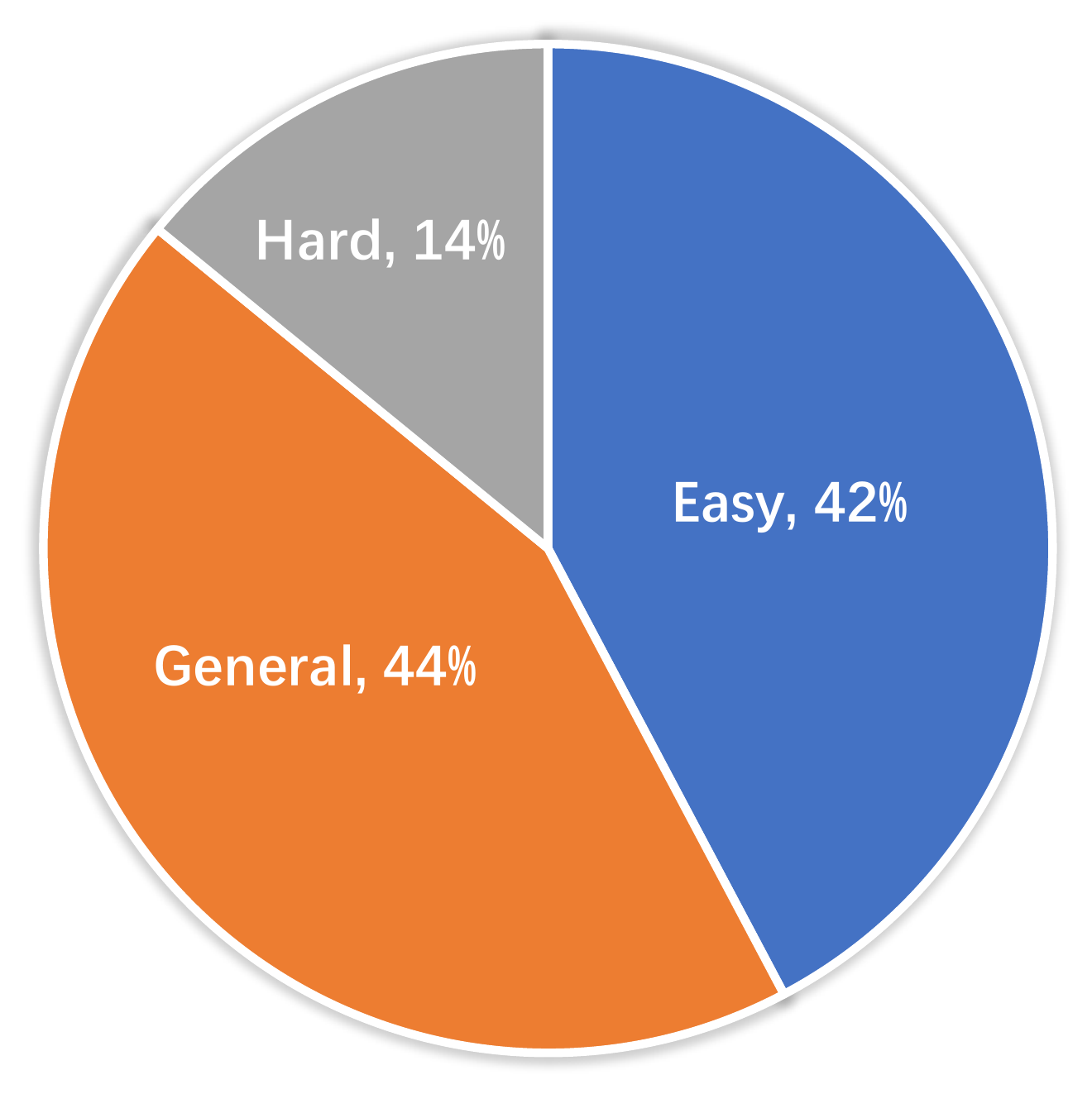}
\caption{Distribution of three types.}
\label{fig-detail} 
\end{figure}

\begin{figure*}[t] 
\centering 
\includegraphics[scale=0.50]{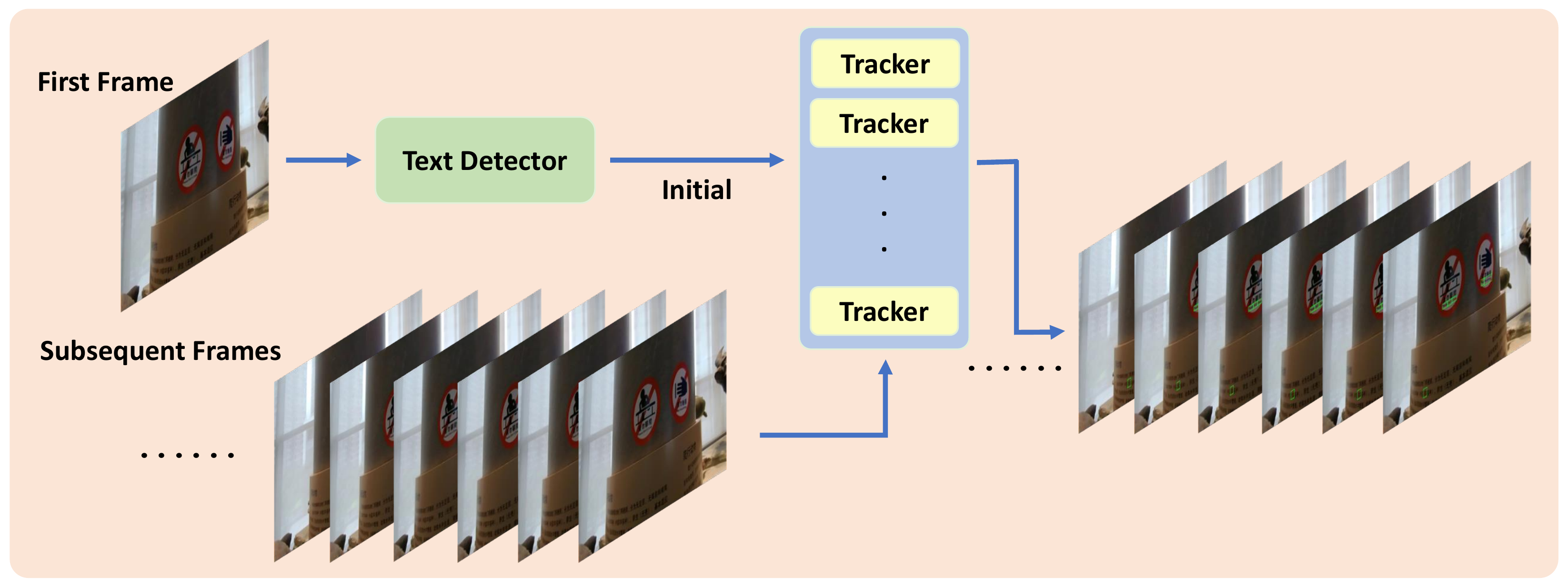}
\caption{Overview of the proposed model for the video scene text detection task. The model consisting of a text detector and multi trackers takes a video (a sequence of frames) as input. The model puts the first frame into the text detector, and use the results to initialize the multi trackers, which are used for tracking the subsequent frames. Best viewed in color and zoom. }
\label{fig-over} 
\end{figure*}

\subsection{annotation}
As described in Section \uppercase\expandafter{\romannumeral4}-B, Text-RBL consists of paired videos including raw videos, blurry videos, and low-resolution videos. For raw videos, we use the proposed semi-automatic labeling strategy, as shown in FIGURE \ref{fig-anno}, while we directly use the annotation of the raw videos to provide the corresponding annotation for paired blurry videos and low-resolution videos without additional operations. Compared with other manually labeled datasets, such as STVText4, which invites over 200 people to conduct three rounds of cross-checking for more than two months, we complete the annotation of Text-RBL in only about 72 hours with the help of the proposed semi-automatic labeling strategy, which greatly reduces the cost of annotation. In fact, it is not difficult for us to further expand Text-RBL in the future.

\subsection{dataset analysis}
Text-RBL contains 369 videos include 101,019 frames and 371,955 different text instances totally, with an average video length of 265 frames. The shortest video contains 34 frames, while the longest one consists of 1,245 frames. Moreover, the number of text instances in each video is different from 1 to 23. Statistics on the full dataset are provided in Table \ref{t4}. 

To investigate the difficulty of Text-RBL in more detail, according to the number of text instances in each video, we categorize Text-RBL into three types. The types of `easy' denotes the number of text instances in videos is below to 4. The types of `general' denotes the number of text instances in videos is between 4 to 8. When the number of text instances in videos is more than 8, these videos is defined as `hard'. As shown in FIGURE \ref{fig-detail}, `easy' and `general'  make up 42\% and 44\% of the dataset respectively, and `hard' makes up the remaining 14\%. Overall, Text-RBL is not too difficult, and more difficult samples can be collected to enrich the dataset in the future.

\section{Baseline Model}

In this section, we introduce our baseline model to verify the rationality of the Text-RBL dataset and evaluate the effectiveness of different combinations of choices of the text detectors and trackers. An overview of the model is illustrated in FIGURE \ref{fig-over}. The model consists of two components, including the text detector and the tracker. For the first frame, we use the detection results performed by the text detector to initialize the trackers whose number is the same as the detection results. For the subsequent frames, we use the initialized multiple trackers to track the text instances to complete the video scene text detection task.

\subsection{Text Detection}
Given a scene text video $\{F_i\}_{i=1}^{n_{frame}}$, where $n_{frame}$ is the length of the video, we first detect the text instances in the first frame $F_1$ by the text detector D, and obtain the detection results as $\{BBox_i^1\}_{i=1}^{n_{text}}$, where $n_{text}$ is the number of text instances detection results. The text detection process can be mathematically formulated as:
\begin{equation}
\{BBox_i^1\}_{i=1}^{n_{text}} = D({F_1})
\end{equation}

Since the detection results of the first frame are used to initialize the tracker, that is to say, the detection results should be as good as possible. Under the consideration of this, We experiment with three representative types of text detector: PSENet \cite{33}, DBNet \cite{34} and Jiang et al. \cite{35}. For these three different text detectors, we do not directly use the pre-trained base-model in the experiment but fine-tuned it on our Text-RBL dataset.

\begin{algorithm}
	\caption{Inference algorithm of the baseline model for video scene text detection.}
	\label{alg:1}
	\begin{algorithmic}[1]
		\Require : Text Detector: $D$; Tracker: $T$
		\State Input: videos frames $\{F_t\}$ with length $l$
		\For{ $t=1$ to $l$}
            \If{$t==1$}
            \State $\{BBox_i^1\}_{i=1}^{n_{text}} = D(F_1)$
            \For{$i=1$ to $n_{text}$}
            \State $T_i = Init(BBox_i^1)$
            \EndFor
            \Else
            \For{$i=1$ to $n_{text}$}
            \State $BBox_i^t = T_i(F_t)$
            \State Update $T_i$
            \EndFor
            \EndIf
        \EndFor
		\State Output: detection and tracking results $\{BBox_i^t\}$
	\end{algorithmic}
\end{algorithm}

\subsection{Text Tracking}
In addition to using the text detector to get the location of the text instances in the first frame of the given video, we use the multi trackers to track text instances in the subsequent frames. 

Normally, the tracking process is that given the target state in the first frame of a video, the tracker predicts the target state in each subsequent frame. Therefore, for each text instance detected in the first frame, we initialize the corresponding tracker with the detection result. Since the detection result is obtained by the text detection method based on segmentation, we need to initialize the tracker with the features in the smallest bounding rectangle of the detection result. For the subsequent frames, the tracking process of the text instances can be mathematically formulated as:
\begin{equation}
BBox_i^t = T_i(F_t)
\end{equation}
where $T_i$ denotes the tracker corresponding to the $i$-$th$ text instance from the top left in the first frame. $F_t$ denotes the frame at time $t$. $BBox_i^t$ denotes the tracking result of the text instance i in the $t$-$th$ frame.

Obviously, the performance of the tracker directly affects the results of the video scene text detection task. Considering the tracking speed and accuracy of the tracker, combined with the simple characteristics of text instances, we use the tracker based on a correlation filter instead of the tracker based on deep learning. We experiment with two correlation filter-based tracker: KCF \cite{11}, and SAMF \cite{13}. More experimental details will later be discussed in Section \uppercase\expandafter{\romannumeral6}.

\subsection{Failure Detection Scheme}

To conquer the baseline model drift problem caused by complex scene text videos (e.g., out-of-view, occlusion, and so on), a simple and effective failure detection scheme is proposed in our work, which can stop the loss in time and improve the performance further.

As the location of the maximum value of the response map indicates the tracking target position, the quality of the final tracking result can be evaluated by the response map. Let $M_R^t$ denotes the maximum value of the response $R$ at $t$-$th$ frame. The maximum response value of the current frame $M_R^t$ and the difference $\Delta{M_R}$ between the maximum response value of the tracking first frame $M_R^1$ are combined to define the confidence score $S_c$ for the current frame, which can be mathematically formulated as:
\begin{equation}
\Delta{M_R} = M_R^t - M_R^1
\end{equation}
\begin{equation}
S_c=\left\{
\begin{array}{rcl}
0,       &      & {M_R^t < \alpha} \ and \  {\Delta M_R < \beta}\\
1,       &      & {otherwise}
\end{array} \right. 
\end{equation}
where $\alpha = 0.25$ and $\beta = -0.2$. The principles of equation (6) include: (1) When $M_R^t$ is not confident and is much lower than $M_R^1$, the tracker outputs a confidence score of 0 and stop tracking the text instance in subsequent frames; (2) Otherwise, the tracker returns a confidence score of 1, which means is conducting successful tracking process.

\section{Experimental Results}
In this section, we first briefly introduce the Text-RBL dataset for training, validation, and testing in Section \uppercase\expandafter{\romannumeral6}-A. Then, we explain the evaluation protocols and introduce the compared baseline models in Section \uppercase\expandafter{\romannumeral6}-B and Section \uppercase\expandafter{\romannumeral6}-C, respectively. 
Finally, we show the comparison results on our Text-RBL dataset for performance evaluation with the effectiveness analysis in Section \uppercase\expandafter{\romannumeral6}-D.

\subsection{Datasets}
In all the experiments, the Text-RBL dataset is divided into three parts: 80\% for training, 10\% for validation, and 10\% for testing. Whether it’s for training, validation, or testing, each part of the Text-RBL dataset consists of paired raw videos, blurry videos, and low-resolution videos.

\subsection{Evaluation protocols}

As mentioned in common text detection algorithms \cite{38}, we evaluate baseline models on our proposed Text-RBL dataset in terms of Precision (denoted by P), Recall (denoted by R),  and F-measure (denoted by F). Precision represents the ratio of the number of correctly detected text instances to the number of all detected text instances. Recall represents the ratio of the number of correctly detected text instances to the number of all text instances in the test dataset.  F-measure measures the overall performance of text detectors, including Recall and Precision. Note that, a correctly detected text instance denotes the overlap between the detected text region and the ground truth of the detected text instance is larger than a given threshold (e.g., 0.5).


\subsection{Compared models}
We define the following combinations of our proposed baseline models for video scene text detection, to evaluate the importance of different text detectors and trackers:

\subsubsection{Detection only}
We use three representative types of text detector in each frame for evaluation: PSENet \cite{33}, DBNet \cite{34} and Jiang et al. \cite{35}. Specifically, PSENet can precisely detect text instances with arbitrary shapes, even can accurately separate text instances that are close to each other with a progressive scale expansion algorithm; DBNet proposes a Differentiable Binarization module to provide a highly robust binarization map, predicting the shrunk regions and significantly simplifying the post-processing; Jiang et al. propose that not only the novel text region representation can detect dense adjacent text of arbitrary shape flexibly, but also the proposed Polygon Expansion Algorithm can reduce the cost of time greatly which needs only one clean and efficient step. For different text detectors, we can define as: \textbf{PSENet only}, \textbf{DBNet only}, and \textbf{Jiang et al. only}.
\subsubsection{Detection+tracking}
For each tested video, we use the text detectors in the first frame and use the trackers in the subsequent frames, which is the full setting baseline model for evaluation. The text detectors are the same as mentioned in Section \uppercase\expandafter{\romannumeral6}-C. The trackers are  KCF \cite{11}, and SAMF \cite{13}. Specifically, KCF is the first one to propose that using the diagonalization property of the circulant matrix in the Fourier space, the calculation of the matrix is transformed into the Hadamard product of the vector, which greatly reduces the amount of calculation, improves the calculation speed, and makes the algorithm meet the real-time requirements. SAMF is a variant of KCF, focusing on suggesting an effective scale adaptive scheme to tackle the problem of the fixed template size. For different combinations of text detectors and trackers, we can define as: \textbf{PSENet+KCF}, \textbf{PSENet+SAMF}, \textbf{DBNet+KCF}, \textbf{DBNet+SAMF}, \textbf{Jiang et al.+KCF}, and \textbf{Jiang et al.+SAMF}.

\subsection{Performance comparison}
\subsubsection{Annotation Comparison}

\begin{figure*}[t] 
\centering 
\includegraphics[scale=0.4]{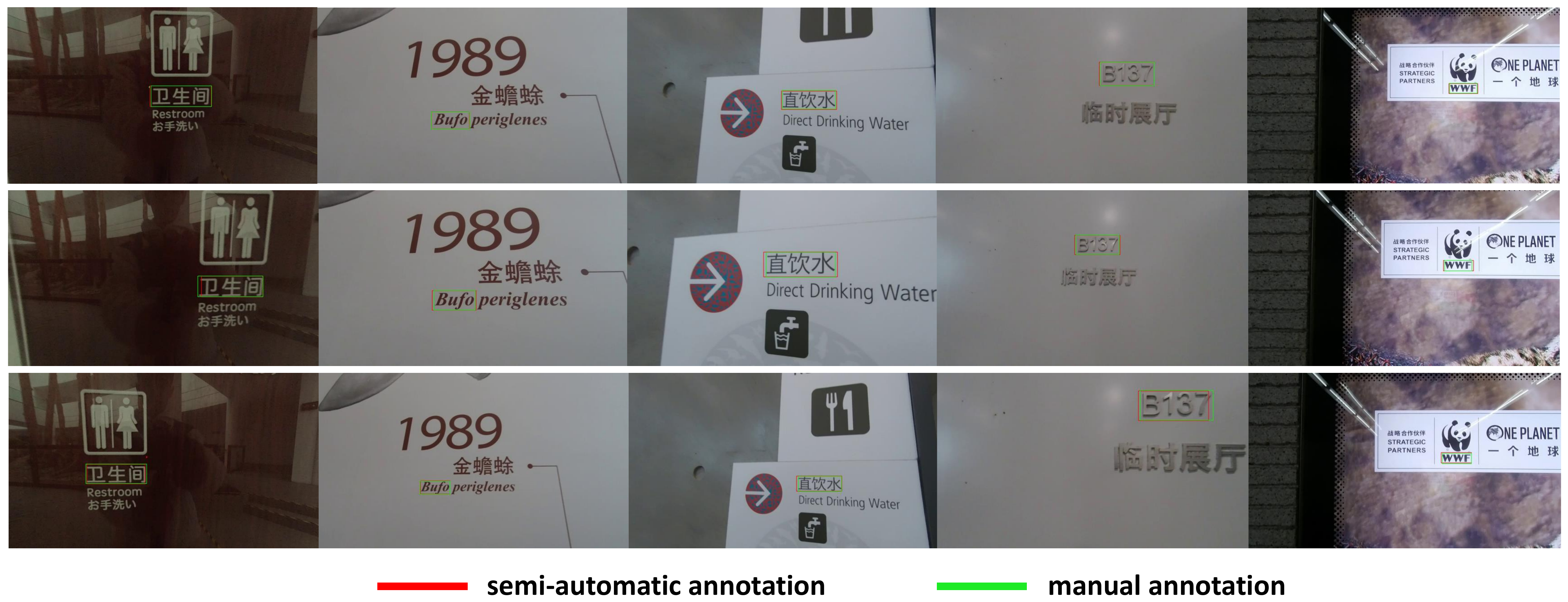}
\caption{Quantitative comparison between manual annotation and semi-automatic text annotation. Best viewed in color and zoom.
\label{fig-munual}}
\end{figure*}

\begin{table}[t]
\centering
\renewcommand\arraystretch{1.3}
\setlength{\tabcolsep}{8.4mm}
\smallskip\begin{tabular}{|l|l|}
\hline
Data & mIOU [\%] \\
\hline
\hline
Video-011 & 92.1 \\
\hline
Video-060 & 88.5 \\
\hline
Video-061  & 89.7 \\
\hline
Video-070 & 91.6 \\
\hline
Video-202 & 93.0 \\
\hline
\end{tabular}
\caption{Quantitative comparison between manual annotation and semi-automatic text annotation.}
\label{t5}
\end{table}

\begin{table*}[t]
\centering
\renewcommand\arraystretch{1.3}
\smallskip\begin{tabular}{|l|l|l|l|l|l|l|l|l|l|l|}
\hline
\multirow{2}*{Method} &\multirow{2}*{Train datatset}& \multicolumn{3}{c|}{Raw-1st} & \multicolumn{3}{c|}{Blurry-1st}  & \multicolumn{3}{c|}{Low-resolution-1st} \\
\cline{3-11}
 & & P [\%] & R [\%] & F [\%] & P [\%] & R [\%] & F [\%] & P [\%] & R [\%] & F [\%]\\
\hline
\hline
\multirow{4}*{PSENet \cite{33}}
 & Raw videos & 87.05 & \textbf{93.19} & 90.01 & 82.18 & 77.25 & 79.64 & 80.85 & 92.46 & 86.27  \\
\cline{2-11}
 & Blurry videos & 91.26 & 91.48 & 91.37 & 90.51 & 88.25 & 89.37 & 86.59 & 86.37 & 86.48 \\
\cline{2-11}
 & Low-resolution videos & 89.69 & 91.00 & 90.34 & 89.33 & 73.25 & 80.49 & 86.97 & 89.29 & 88.12 
\\
\cline{2-11}
 & Mix videos & \textbf{93.19} & \textbf{93.19} & \textbf{93.19} & \textbf{93.80} & \textbf{90.75} & \textbf{92.25} & \textbf{92.84} & \textbf{91.48} & 
 \textbf{92.16}
\\
\hline
\hline
\multirow{4}*{DBNet \cite{34}}
  & Raw videos & 91.73 & 83.70 & 87.53 & 89.52 & 70.50 & 78.88 & 91.04 & 76.64 & 83.22 
 \\
\cline{2-11}
& Blurry videos & 93.44 & 83.21 & 88.03 & 92.12 & 76.00 & 83.29 & 91.92 & 80.29 & 85.71 
 \\
\cline{2-11}
& Low-resolution videos & 95.39 & 85.64 & 90.26 & \textbf{94.50} & 73.00 & 82.37 & \textbf{96.66} & \textbf{84.43} & \textbf{90.13}
\\
\cline{2-11}
 & Mix videos & \textbf{96.71} & \textbf{85.89} & \textbf{90.98} & 91.90 & \textbf{82.25} & \textbf{86.81} & 96.39 & \textbf{84.43} & 90.01 
\\
\hline
\hline
\multirow{4}*{Jiang et al. \cite{35}}
  & Raw videos & 93.15 & 89.29 & 91.18 & 86.96 & 65.00 & 74.39 & 87.60 & 79.08 & 83.12  \\
\cline{2-11}
 & Blurry videos & \textbf{96.61} & 90.27 & 93.33 & 93.36 & 81.25 & 86.90 & \textbf{96.43} & 85.40 & 90.58 \\
\cline{2-11}
& Low-resolution videos & 94.64 & 90.27 & 92.40 & 91.30 & 68.25 & 78.11 & 91.62 & 87.83 & 89.69 
 \\
\cline{2-11}
 & Mix videos & 96.20 & \textbf{92.46} & \textbf{94.29} & \textbf{95.47} & \textbf{89.50} & \textbf{92.39} & 94.22 & \textbf{91.24} & \textbf{92.71}\\
\hline
\end{tabular}
\caption{Performance comparison results of different trackers trained with different datasets on different testing datasets. $P$, $R$ and  $F$ represent the precision, recall, and F-measure, respectively.}
\label{t1}
\end{table*}
\begin{table*}[t]
\centering
\renewcommand\arraystretch{1.3}
\smallskip\begin{tabular}{|l|l|l|l|l|l|l|l|}
\hline
\multirow{2}*{Method} &\multirow{2}*{Train datatset}& \multicolumn{3}{c|}{IC15 \cite{3}} & \multicolumn{3}{c|}{Raw-1st (Text-RBL)} \\
\cline{3-8}
 & & P [\%] & R [\%] & F [\%] & P [\%] & R [\%] & F [\%] \\
\hline
\hline
\multirow{2}*{Jiang et al. \cite{35}}
  & IC15 \cite{3} & 77.79 & \textbf{74.53} & 76.12 & 59.62 & 60.34 & 59.98  \\
\cline{2-8}
 & IC15 \cite{3} + Raw videos (Text-RBL) & \textbf{78.22} & \textbf{74.53} & \textbf{76.33} & \textbf{94.62} & \textbf{89.78} & \textbf{92.13}  \\
\hline
\end{tabular}
\caption{Performance comparison results with IC15\cite{3}. $P$, $R$ and $F$ represent the precision, recall, and F-measure, respectively.}
\label{t7}
\end{table*}

\begin{table*}[t]
\centering
\renewcommand\arraystretch{1.3}
\smallskip\begin{tabular}{|l|l|l|l|l|l|l|l|l|l|l|}
\hline
\multirow{2}*{Detector} &\multirow{2}*{Tracker}& \multicolumn{3}{c|}{Raw-all} & \multicolumn{3}{c|}{Blurry-all}  & \multicolumn{3}{c|}{Low-resolution-all} \\
\cline{3-11}
 & & P [\%] & R [\%] & F [\%] & P [\%] & R [\%] & F [\%] & P [\%] & R [\%] & F [\%]\\
\hline
\hline
\multirow{3}*{PSENet \cite{33}}
& -  & 92.75 & 93.25 & 93.00 & 92.30 & 90.86 & 91.58 & 93.75 & 92.74 & 93.24 \\
\cline{2-11}
& KCF \cite{11} & 90.49 & 89.37 & 89.93 & 90.38 & 93.27 & 91.80 & 94.40 & 93.75 & 94.07\\
\cline{2-11}
& SAMF \cite{13} & \textbf{98.75} & \textbf{97.53} & \textbf{98.14} & \textbf{94.17} & \textbf{97.18} & \textbf{95.65} & \textbf{96.96} & \textbf{96.59} & \textbf{96.77}
\\
\hline
\hline
\multirow{3}*{DBNet \cite{34}}
 & - & 96.82 & 86.72 & 91.49 & 92.60 & 81.46 & 86.67 & 97.05 & 86.19 & 90.73 \\
\cline{2-11}
& KCF \cite{11} & 95.21 & 88.58 & 91.77 & 95.04 & 88.28 & 91.53 & 94.71 & 92.98 & 93.84\\
\cline{2-11}
& SAMF \cite{13} & \textbf{99.12} & \textbf{92.22} & \textbf{95.55} & \textbf{95.95} & \textbf{89.12} & \textbf{92.41} & \textbf{94.92} & \textbf{93.19} & \textbf{94.05}\\
\hline
\hline
\multirow{3}*{Jiang et al. \cite{35}}
& -  & 96.78 & 92.85 & 94.77 & \textbf{95.05} & 89.30 & 92.09 & \textbf{95.11} & 91.49 & 93.26 \\
\cline{2-11}
& KCF \cite{11} & 93.21 & 90.03 & 91.59 & 93.57 & 91.16 & 92.53 & 94.20 & 94.45 & 94.33 
\\
\cline{2-11}
& SAMF \cite{13} & \textbf{97.23} & \textbf{93.91} & \textbf{95.54} & 93.95 & \textbf{91.53} & \textbf{92.73} & 94.60 & \textbf{94.85} & \textbf{94.73}
\\
\hline
\end{tabular}
\caption{Performance comparison results of different baselines trained with mix videos on different testing datasets. $P$, $R$ and  $F$ represent the precision, recall, and F-measure, respectively.}
\label{t2}
\end{table*}

\begin{table*}[t]
\centering
\renewcommand\arraystretch{1.3}
\smallskip\begin{tabular}{|l|l|l|l|l|}
\hline
\multirow{2}*{Detector} &\multirow{2}*{Tracker}& \multicolumn{3}{c|}{Raw-all}\\
\cline{3-5}
 & & P [\%] & R [\%] & F [\%]\\
\hline
\hline
\multirow{2}*{PSENet \cite{33}}
& KCF w/o failure detection scheme & 90.49 & \textbf{89.37} & 89.93 \\
\cline{2-5}
& KCF w/ failure detection scheme  & \textbf{95.65} & 89.35 & \textbf{92.39} \\
\hline
\hline
\multirow{2}*{DBNet \cite{34}}
& KCF w/o failure detection scheme & 95.21 & \textbf{88.58} & 91.77 \\
\cline{2-5}
& KCF w/ failure detection scheme & \textbf{97.75} & \textbf{88.58} & \textbf{92.94} \\
\hline
\hline
\multirow{2}*{Jiang et al. \cite{35}}
& KCF w/o failure detection scheme & 93.21 & \textbf{90.03} & 91.57\\
\cline{2-5}
& KCF w/ failure detection scheme  & \textbf{95.73} & 89.87 & \textbf{92.71}\\
\hline
\end{tabular}
\caption{Performance comparison results on Text-RBL with different failure detection scheme. $P$, $R$ and  $F$ represent the precision, recall, and F-measure, respectively.}
\label{t6}
\end{table*}

\begin{figure*}[t] 
\centering 
\includegraphics[scale=0.4]{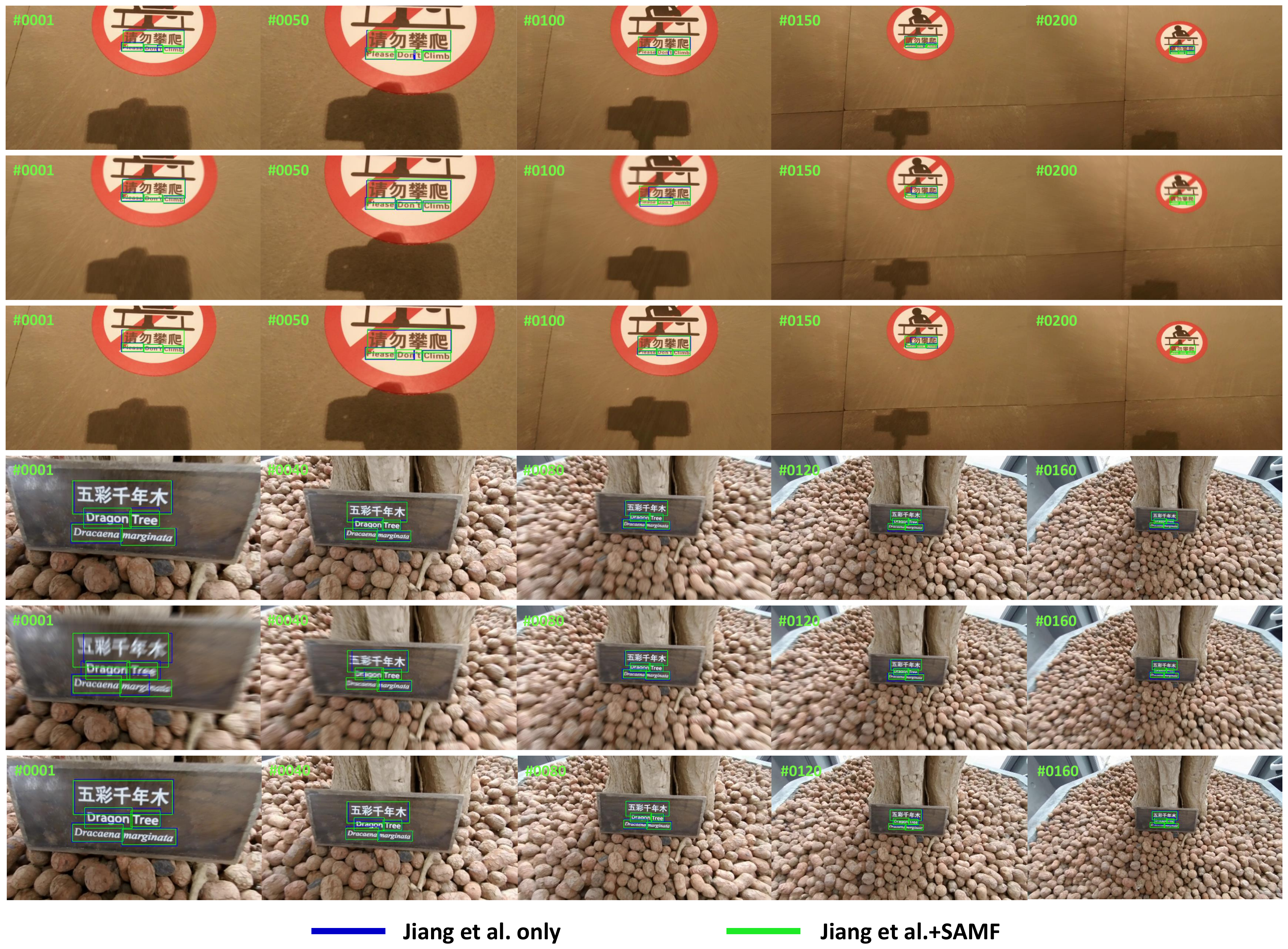}
\caption{ Qualitative  evaluation  of  \textbf{Jiang et al. only} trained by raw vidoes and \textbf{Jiang et al.+SAMF} trained by mix videos on Text-RBL. \textbf{Jiang et al.+SAMF} performs well against  \textbf{Jiang et al. only}. Best viewed in color and zoom.}
\label{fig-quality} 
\end{figure*}

To prove that the KCF tracker used in the semi-automatic text annotation strategy is sufficiently robust and stable, a comparative experiment of manual annotation and semi-automatic text annotation is provided. The evaluation metric is the calculation of mIOU, which is commonly used in the field of detection and tracking. 

In order to ensure the fairness and effectiveness of the experiment, we randomly select five different scene text instances from five different scene text videos for manual annotation, which contains 539 frames in total. The corresponding semi-automatic annotation is obtained by the method mentioned in Section \uppercase\expandafter{\romannumeral3}.

As shown in Table \ref{t5}, regardless of any scene text instance, the results of manual annotation and the proposed semi-automatic annotation are roughly the same, and their mIOU are very high, some even as high as 93\%. This is enough to prove that the tracking based semi-automatic annotation for scene text videos is reliable. Moreover, to further analyze the gap between manual annotation and semi-automatic annotation, the results of the both annotation are visualized, as shown in FIGURE \ref{fig-munual}. Obviously, any scene text instance can get a good annotation result, whether it is manually annotated or the tracking based semi-automatic annotation strategy. The slight difference is that the semi-automatic annotation results may be slightly off the boundary of the scene text instances, but this will not cause a big adverse effect. Through qualitative and quantitative experiments, we have reasons to believe that the tracking based semi-automatic annotation strategy is trustworthy.

\subsubsection{Text Detector Comparison}
Whether it is \textbf{detection only} or  \textbf{detection+tracking}, the text detectors play an extremely important role in video scene text detection. In order to compare the performance of different text detectors on our dataset concisely and clearly, Table \ref{t1} shows the performance of three text detectors used in the baseline models on different types of the testing dataset including raw videos, blurry videos, and low-resolution videos, which are trained with different types of training dataset also including raw videos, blurry videos, low-resolution videos, and mix videos. Mix videos are a mixture of raw videos, blurry videos, and low-resolution videos, but their numbers remain the same. Note that, the testing dataset here only contains the first frame of each video instead of the entire video, defined as raw-1st, blurry-1st, and low-resolution-1st. 

As shown in Table \ref{t1}, although the models trained only on raw videos performs well on the raw-1st, the performance on the blurry-1st and low-resolution-1st is not satisfactory. Notably, the F-measure of PSENet on the raw-1st is 90.01\%, however, its F-measure on the low-resolution-1st drops to 86.27\%, and its F-measure on the raw-1st is only 79.64\%.  However, the models trained with challenging training data such as blurry videos and low-resolution videos can significantly improve performance on blurry-1st and low-resolution-1st. After training with mixed videos, the  F-measure of PSENet boost to 93.19\%,  92.25\%, and 92.16\%, on the raw-1st, blurry-1st, and low-resolution-1st  respectively. In addition to PSENet, DBNet and Jiang et al. also have similar performance improvements. The specific experimental results are shown in Table \ref{t1}.

Table \ref{t7} shows the performance gain of Jiang et al. brought by Text-RBL compared with the previous dataset IC15. Especially on raw-1st, Jiang et al. trained by IC15 and raw videos of Text-RBL get large performance gains in  precision (+35.0\%), recall (+29.44\%) and F-measure (+32.15\%), compared with the one only trained by IC15.

Fundamentally, we observe that it is necessary to propose such a challenging video scene text dataset, which can improve the robustness of the model and enable the model to better deal with challenges such as low-resolution and blurring in real scenes.

\subsubsection{Baseline Comparison}
For the video scene text video task, we use the evaluation protocol to indicate the performance of each baseline with different combinations. To facilitate the distinction, we define the testing datasets here as raw-all, blurry-all, and low-resolution-all. 

Table \ref{t2} summarizes the main results, and all text detectors in the table are trained on the mix videos. For the same text detector, such as DBNet, \textbf{DBNet only}, \textbf{DBNet+KCF}  and \textbf{DBNet+SAMF} show completely different performance, especially on challenging datasets, \textbf{DBNet+KCF} and \textbf{DBNet+SAMF}  significantly advances \textbf{DBNet only} by large margins in F-measure (+4.86\% and +5.74\%, respectively) on the blurry-all. Similar conclusions can be obtained on PSENet and Jiang et al. Particularly, no matter which testing dataset or evaluation protocols is, compared with \textbf{PSENet only}, the performance of \textbf{PSENet+SAMF} is very competitive. The specific experimental results are shown in Table \ref{t2}. We analyze that compared with the text detector itself, the tracker can use the results of the first frame of text detection to better handle the deformation, blur, and low resolution of the text instances in the subsequent frames. 

To validate the effectiveness of the proposed failure detection scheme, we implement a control method by disabling this scheme. The compared results are shown in Table \ref{t6} . The methods with the failure detection scheme obtain large increases in precision and F-measure respectively compared with the methods without scheme. Although \textbf{PSENet+KCF} with failure detection scheme drops 0.02\% in recall, it obtains gains of +5.16\% and +2.46\% in precision and F-measure over the one without failure detection scheme respectively. In addition to PSENet, DBNet and Jiang et al. also have similar performance improvements. The specific experimental results are shown in Table \ref{t6}.

\subsubsection{Speed Comparison}
Although the baselines we proposed are all two-stage combined with text detection and tracking, they not only enhance the performance of video scene text detection but also greatly improves speed, which benefits from the real-time tracking performance of the tracker. The results of the speed performance comparison between different baselines are presented in Table \ref{t3}. Benefit from the fast tracking performance of KCF, \textbf{Jiang et al.+KCF} achieves a real-time speed of about 106 FPS, while the speed of  \textbf{Jiang et al.+SAMF} is 5.38 FPS, and \textbf{Jiang et al. only} is only 2.73 FPS.

\begin{table}[t]
\centering
\renewcommand\arraystretch{1.3}
\setlength{\tabcolsep}{8.4mm}
\smallskip\begin{tabular}{|l|l|}
\hline
Method & Speed \\
\hline
\hline
\textbf{PSENet only} & 0.71 fps \\
\hline
\textbf{DBNet only} & 4.20 fps \\
\hline
\textbf{Jiang et al. only}  & 2.73 fps \\
\hline
\textbf{Jiang et al.+KCF} & 106 fps \\
\hline
\textbf{Jiang et al.+SAMF} & 5.38 fps \\
\hline
\end{tabular}
\caption{Speed compared on different baselines for video scene text detection.}
\label{t3}
\end{table}

\subsubsection{Qualitative Analysis}
FIGURE \ref{fig-quality} illustrates some qualitative results of \textbf{Jiang et al. only} trained by raw videos and \textbf{Jiang et al.+SAMF} trained by mix videos on Text-RBL. For each video, we present the detection results of the raw video, blurry video, and low-resolution video from top to bottom. We can find that the baseline model \textbf{Jiang et al.+SAMF} trained by mix videos on Text-RBL performs better on challenging data compared with \textbf{Jiang et al. only} trained by raw videos. On some low-quality frames, \textbf{Jiang et al. only} even can not detect any scene text instances, as shown in FIGURE \ref{fig-quality}. This once again demonstrates the validity of the proposed Text-RBL dataset.

\section{Future Work}
Benefiting from our proposed semi-automatic labeling strategy and paired low-quality scene text video generation method, we can easily obtain a large number of low-quality scene text videos with annotations in the future. A large number of low-quality scene text videos with annotations are very useful and urgently needed in the field of text recognition. We can use paired videos to improve the performance of the text recognition network.

\section{Conclusion}
In this paper, in order to release the human, material, and financial resources brought by the annotation of scene text videos, a semi-automatic labeling strategy based on tracking is proposed. The text instances in the first frame are manually labeled, which are used to initialize the trackers, and the trackers automatically label the text instances in the subsequent frames. Moreover, a novel and challenging scene text video dataset is proposed, consisting of raw videos, blurry videos, and low-resolution videos, named Text-RBL. In addition, we propose a baseline combining the text detection and tracking into a unified framework with a failure detection scheme for video scene text detection. The extensive experiments on Text-RBL demonstrate that baselines especially those trained on low-quality videos perform favorably against detection only methods not only on accuracy but also on speed. It is necessary and urgent to propose such a challenging scene text video dataset with effective annotation.

\bibliographystyle{IEEEtran}
\bibliography{ref}

\EOD

\end{document}